# SKULL-STRIPPING FOR TUMOR-BEARING BRAIN IMAGES

**Stefan Bauer, Lutz-P. Nolte, Mauricio Reyes**

*Institute for Surgical Technology and Biomechanics, University of Bern, Switzerland*

## Introduction

Skull-stripping separates the skull region of the head from the soft brain tissues. In many cases of brain image analysis, this is an essential preprocessing step in order to improve the final result. This is true for both registration and segmentation tasks. In fact, skull-stripping of magnetic resonance images (MRI) is a well-studied problem with numerous publications in recent years. Many different algorithms have been proposed, a summary and comparison of which can be found in [Fennema-Notestine, 2006]. Despite the abundance of approaches, we discovered that the algorithms which had been suggested so far, perform poorly when dealing with tumor-bearing brain images. This is mostly due to additional difficulties in separating the brain from the skull in this case, especially when the lesion is located very close to the skull border. Additionally, images acquired according to standard clinical protocols, often exhibit anisotropic resolution and only partial coverage, which further complicates the task. Therefore, we developed a method which is dedicated to skull-stripping for clinically acquired tumor-bearing brain images.

## Methods

We adopted a two-step procedure: In a first step, a standard brain atlas [Talos, 2010] is registered to the patient image with an affine registration and the brain mask of this atlas is propagated using the calculated transformation matrix. In order to speed-up the algorithm, the registration step is performed on a subsampled version of the original image. The transformed brain mask serves as an initialization for a level-set based refinement of the brain region in the second step. We use a geodesic active contours level-set method with dedicated balloon force, curvature force and advection force, which evolves towards the brain-skull border. The initialization with an affine registration makes the whole procedure very robust because the level-set segmentation has to account only for small deformations around the border of the skull.

## Results

The method has been evaluated on a publicly available dataset of healthy MR images of the head [Shattuck, 2009] and on tumor-bearing brain images from the ContraCancrum database [Marias, 2011]. For the standard dataset, quantitative evaluation was done using Dice coefficient. The Dice coefficient measures the overlap with the ground-truth segmentations. It can range between 0 and 1, with 1 indicating perfect overlap. On 40 healthy datasets we achieved a Dice coefficient of $0.86 \pm 0.03$. The brain tumor images from the ContraCancrum database could only be evaluated qualitatively. Visual inspection showed convincing results as demonstrated for one case in figure 1. Computation time for the method ranged between 2 and 3 minutes on a standard PC, depending on the size of the dataset.

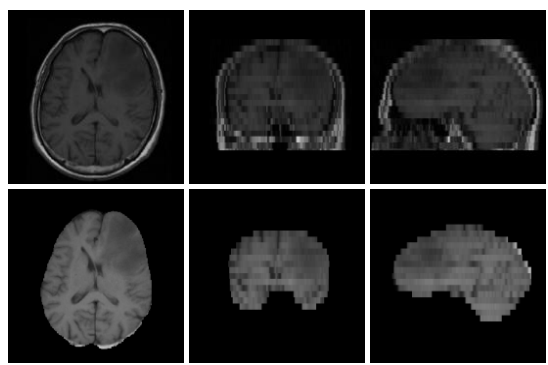

*Figure 1: One volumetric tumor-bearing brain image before and after skull-stripping in axial, coronal and sagittal view*

## Discussion & Conclusion [1]

We presented a robust and fully automatic method for skull-stripping, which achieves better results than the standard methods on tumor-bearing brain images. Computation time is reasonably fast on 3D MRI volumes. The tool is publicly available [2] and it has also been integrated as a plugin into the DoctorEye software platform [Marias, 2011].

[1] Funding by the European Union within the framework of the ContraCancrum project (FP7 - IST-223979) is gratefully acknowledged.

[2] http://www.istb.unibe.ch/content/surgical_technologies/medical_image_analysis/software/